\documentclass[CGV,biber,table]{nowfnt} 

\usepackage[utf8]{inputenc}
\usepackage{amssymb}
\usepackage{lscape}
\usepackage{colortbl}
\usepackage{textcomp}
\usepackage{booktabs}
\newcommand{\ie}{\emph{i.e.}}
\newcommand{\eg}{\emph{e.g.}}
\newcommand{\cf}{\emph{c.f.}}
\newcommand{\etc}{\emph{etc}}

\newcommand*\rot{\rotatebox{90}}

\definecolor{lightgray}{gray}{0.9}

\title{Video Summarization Overview}

\maintitleauthorlist{
Mayu Otani \\
CyberAgent, Inc. \\
otani\_mayu@cyberagent.co.jp
\and
Yale Song \\
Microsoft Research \\
yalesong@microsoft.com
\and
Yang Wang \\
University of Manitoba \\
ywang@cs.umanitoba.ca
}

\issuesetup
{%
 copyrightowner={M.~Otani and Y.~Song and Y.~Wang},
 volume        = 13,
 issue         = 4,
 pubyear       = 2022,
 doi           = 10.1561/0600000099,
 firstpage     = 284, 
 lastpage      = 335
 }

\usepackage{mwe}

\author[1]{Otani,Mayu}
\author[2]{Song,Yale}
\author[3]{Wang,Yang}

\affil[1]{CyberAgent, Inc.; otani\_mayu@cyberagent.co.jp}
\affil[2]{Microsoft Research; yalesong@microsoft.com}
\affil[3]{University of Manitoba; ywang@cs.umanitoba.ca}

\articledatabox{\nowfntstandardcitation}

\begin{document}

\makeabstracttitle

\begin{abstract}
With the broad growth of video capturing devices and applications on the web, it is more demanding to provide desired video content for users efficiently. Video summarization facilitates quickly grasping video content by creating a compact summary of videos. Much effort has been devoted to automatic video summarization, and various problem settings and approaches have been proposed. 
Our goal is to provide an overview of this field. This survey covers early studies as well as recent approaches which take advantage of deep learning techniques. We describe video summarization approaches and their underlying concepts. We also discuss benchmarks and evaluations. We overview how prior work addressed evaluation and detail the pros and cons of the evaluation protocols. Last but not least, we discuss open challenges in this field.
\end{abstract}
\chapter{Introduction}

The wide spread use of internet and affordable video capturing devices have dramatically changed the landscape of video creation and consumption.
In particular, user-created videos are more prevalent than ever with the evolution of video streaming services and social networks.
The rapid growth of video creation necessitates advanced technologies that enable efficient consumption of desired video content.
The scenarios include enhancing user experience for viewers on video streaming services, enabling quick video browsing for video creators who need to go through a massive amount of video rushes, and for security teams who need to monitor surveillance videos.
    
Video summarization facilitates quickly grasping video content by creating a compact summary of videos.
One naive way to achieve video summarization would be to increase the playback speed or to sample short segments with uniform intervals.
However, the former degrades the audio quality and distorts the motion~\citep{benaim2020speednet}, while the latter might miss important content due to the random sampling nature of the method.
Rather than these naive solutions, video summarization aims to extract the information desired by viewers for more effective video browsing.

The purpose of video summaries varies considerably depending on application scenarios.
For sports, viewers want to see moments that are critical to the outcome of a game, whereas for surveillance, video summaries need to contain scenes that are unusual and noteworthy.
The application scenarios grow as more videos are created, \eg, we are beginning to see new types of videos such as video game live streaming and video blogs (vlogs).
This has led to a new problem of video summarization as different types of videos have different characteristics and viewers have particular demands for summaries.
Such variety of applications have stimulated heterogeneous research in this field.

Video summarization addresses two principal problems: ``what is the nature of a desirable video summary'' and ``how can we model video content.''
The answers depend on application scenarios.
While these are still open problems for most application scenarios, many promising ideas have been proposed in the literature.
Early work made various assumptions about requirements for video summaries, \eg, uniqueness (less-redundancy), diversity, and interestingness.
Some works focused on creating video summaries that are relevant to user's intention and involve user interactions.
Recent research focuses more on data-driven approaches that from annotated datasets to learn desired video summaries.

Computational modeling of desirable video content is also an important challenge in video summarization.
Starting with low-level features, various feature representations have been applied, such as face recognition and visual saliency.
Recently, feature extraction using deep neural networks has been mainly adopted.
Some applications further utilize auxiliary information such as subtitles for documentary videos, game logs for sports videos, and brain waves for egocentric videos captured with wearable cameras.

The goal of this survey is to provide a comprehensive overview of the video summarization literature. We review various video summarization approaches and compare their underlying concepts and assumptions. We start with early works that proposed seminal concepts for video summarization, and also cover recent data-driven approaches that take advantage of end-to-end deep learning. By categorizing the diverse research in terms of application scenarios and techniques employed, we aim to help researchers and practitioners to build video summarization systems for different purposes and application scenarios.

We also review existing benchmarks and evaluation protocols and discuss the key challenges in evaluating video summarization, which is not straightforward due to the difficulty of obtaining ground truth summaries. We provide an overview of how previous works have addressed challenges around evaluation and discuss strengths and weaknesses of existing evaluation protocols. Finally, we discuss open challenges in this area.
\chapter{Taxonomy of Video Summarization}

We broadly define video summarization as a task of creating a short-form video content from the original long video so that users can quickly grasp the main content of interest. The most typical form of video summarization is video skimming, which selects interesting moments from the original video and puts them together into a short-form video. There also exists a wide variety of other task definitions in the field of video summarization, which we will review in this section. In addition, there is a relevant task called video highlight detection, which finds a short interesting moment in a video. Following our definition, we will also cover such relevant tasks as a sub-domain of video summarization.

Different definitions of video summarization have different purposes and output representations, \eg, creating a video skim of a family event video or creating a natural language summary of a movie. This chapter provides an overview of various video summarization problems.
We begin by describing target domains for video summarization. We explore typical domains and compare their characteristics. We also list typical video summary formats. We expect this section will help readers clarify their own problem setting and find an appropriate summary format for their scenarios.


\section{Video Domains}
A variety of video domains have been studied in video summarization, \eg, documentaries, sports, news, how-to tutorials, and egocentric videos. The domain is an important factor in developing a video summarization system because the definition of a ``good'' summary depends on it. For example, in sports videos, the moment that determines the outcome of a game is important, while for cooking videos, it is important to cover all the necessary procedures.

Also, different domains provide auxiliary information that can be leveraged for video summarization.
Videos in some domains are created for the public audience, including news, sports, and movies. These videos are professionally edited and often come with rich metadata, \eg, movies come with subtitles, scripts, identity of actors, \etc.

Some domains lacks such metadata, especially those mainly for personal purposes, \eg, home videos. These characteristics introduce challenges in video content understanding in general. On the other hand, there is a rise of on-demand video analysis tools including speech recognition, object detection, shot detection, and face recognition~\citep{mediapipe, Poms:2018:Scanner}. These analytic tools help developers and researchers annotate video content.
In the following, we introduce video domains and their characteristics in terms of video summarization.

\paragraph{Movies}
Movie summarization has been a topic of interest since the early days of this area \citep{PFEIFFER1996345}. Typically, the goal here is to generate an engaging summary of the storyline. The output representation for this scenario is often a short video, \eg, a movie trailer, and contains scenes featuring the main characters or those in significant events of a story. In some cases, such as movie trailer generation, a summary has to avoid accidentally revealing critical parts of the storyline. Videos in this domain are often associated with rich metadata. Analyzing subtitles is an effective way to understand a storyline~\citep{Evangelopoulos2013}. In the literature, analyzing metadata such as characters have been explored \citep{Sang2010}. Leveraging such auxiliary information can often help develop a summarization system. As an example that leverages rich movie metadata, SceneSkim~\citep{sceneskim} is a video browser that helps users quickly find scenes from a movie. They achieve this by associating manually created textual plot summary, captions, and scripts with scenes and enable users to search over the metadata.

\paragraph{Broadcast Sports} The typical goal of broadcast sports video summarization is to allow users to watch scenes of significant events or those of favorite players.
In sports video summarization, domain-specific knowledge is a helpful clue, \eg, the rules of a sport game. Leveraging the underlying rules facilitates performing inference and reasoning at a high-level, which is also interpretable to humans, \eg, in American football, touchdowns, field goals, and extra points are significant events that affect the outcome of a game \citep{Babaguchi2004}. By detecting these events, we can obtain highlights of a game. This feature of sports videos can be used not only for summarizing broadcast sports videos, but also for those recorded by non-professional users. Other effective features include text overlays inserted during editing, replays, and audience cheers. 

Another new type of sports video is eSports \citep{Song2016}, which is a video game competition. Unlike other sports videos, the main content of eSports videos is computer graphics game scenes. As eSports is a relatively new domain, there has not been much research yet, \cf, \citep{Song2016, block2018narrative}.

\paragraph{News and Documentaries}
These domains have been included in several video summarization datasets and have received much attention in this field. News and documentaries convey information in both visual and audio narrations. Therefore, summaries need to retain crucial audio content in addition to visual content. \cite{Taskiran2006} proposed to analyze audio transcripts of documentaries to create summaries. In addition, news videos are characterized by the fact that various videos are created for the same event. Therefore, summarizing multiple news videos has also been explored. For example, \cite{Liu2012} use news videos from different communities and summarize them to emphasize the differences in the content that each community is focusing on.


\paragraph{Home Videos}
Home videos are typically recorded by non-professional users, often for the purpose of recording memorable events and activities. Home videos are mostly consumed by families, relatives, and friends. Non-professional users are usually unfamiliar with editing videos and annotating metadata. Thus, it has been considered to be difficult to make assumptions about the underlying structure of home videos. Most methods are based on heuristic-based objectives such as visual diversity, representativeness, or change point detection \citep{Zhao2014}, \etc. After the success of deep neural networks, many methods have been proposed to learn a model to predict importance scores from annotated video summary datasets \citep{zhang16_eccv,10.1007/978-3-030-01258-8_22}.

There are video hosting platforms on internet where a large number of user-generated videos are uploaded. On such platforms, we can find a large amount of videos that cover similar topics of interest, \eg, surfing, diving, changing tires, etc. Several previous studies have explored using such video collections for video summarization \citep{Chu2015,8099938}. These methods are based on an intuition that we may obtain some clues about common patterns in the relevant video collections even if it is difficult to obtain structure from a single user generated video.

\paragraph{Surveillance Videos}
Surveillance video is generally captured by fixed-point cameras. Surveillance cameras can be seen at offices, hospitals, public places, etc. The purpose of surveillance video summarization is to eliminate redundant scenes and to capture interesting and noteworthy events. Surveillance video summarization faces some unique challenges. One is the massive amount of video data due to long recording times. A summarization system has to handle the massive video data efficiently. In addition, videos are hardly annotated with metadata, thus video content must be recognized only with visual cues. Some surveillance videos are produced by a network of multiple cameras. In this video domain, summarization methods for such multi-view videos are also proposed \citep{7934321}.

\paragraph{Egocentric Videos}
Egocentric videos are captured using wearable cameras. The purposes of egocentric videos also vary. Police officers in some countries use wearable cameras to record events encountered during work hours. Some people record special events and life-loggers record their daily lives.
Previous studies addressed video summarization for egocentric videos \citep{Lu2013,Lee2012,Xiong2015,Xu2015,narasimhan2021clip}.
As egocentric videos are associated with users motions, motion features are often used to infer states of the user \citep{Lu2013,5995406}.
Other major egocentric video features are objects, interactions to objects, faces and so on \citep{Lu2013,Lee2012}.

\paragraph{360\textdegree~Videos}
A 360\textdegree video has a field of view that covers almost all directions. Omnidirectional cameras are used to capture 360\textdegree~videos, and the video is usually edited into a spherical video.
Summarizing a 360\textdegree~video requires locating important content temporally and spatially.
Previous studies formulate 360\textdegree~video summarization as a problem of automatically controlling virtual camera motion so that viewers do not have to search where to look by themselves \citep{8099633,pano2vid}. 

\paragraph{Live Streaming Videos}
Live streaming videos are captured and broadcasted without editing, and therefore tend to be highly redundant. Thus, summarizing a live streaming video requires eliminating many redundant scenes and finding the content of interest.
The characteristic of live streaming videos is real-time interactions by viewers, \eg, adding chat messages \citep{10.1145/3375959.3375965}.
Another characteristic is that live streaming often lasts several hours and many viewers join in the middle of the stream. Previous research proposed an application that displays a summary of what has happened in the stream for viewers who have just joined the stream \citep{10.1145/3491102.3517461}.

As we reviewed, video summarization becomes a completely different problem depending on target domain.
For example, we need different solutions to summarize movies vs. sports videos: A movie may have subtitles, but a sports video may not. On the other hand, for sports videos, we know when important events occurred from the game logs, but such metadata may not be available in a movie. Furthermore, the scenes that viewers want are very different between sports videos and movies.
Thus, in order to build a video summarization system, it is important to first clarify what kind of videos the system would take as input and what the viewers is looking for.
By doing so, we can choose an appropriate approach and summary type.

\section{Purposes of Video Summaries}
In the beginning of this section, we have defined that the goal of video summarization is helping the user to grasp the main content of interest. Here we further provide the details of the purpose of video summaries with a few examples.
The purpose of video summaries depends on video domains and applications. We list typical purposes of video summarization as below.

\begin{itemize}
\item \textbf{Present important content in a video to viewers}
Summaries extract only the important content so viewers can get a quick glimpse of a video without having to watch the whole video. In this scenario, the viewer is assumed not necessarily to watch the original video. This type of summary is particularly effective, \eg, for showing a synopsis of a previous episodes of a TV drama series or for highlighting of a sports game. An important challenge here is how the video summarization system determines the importance of the content in a video.

\item \textbf{Locate specific moments of interest}
Such video summaries present an overview of what the video contains and help the viewer find the desired moments. The design of output summary representation is an important challenge for this purpose. In this scenario, we assume that the viewer will go back to the original video after finding where to watch. Such summaries make browsing surveillance videos more effective.

\item \textbf{Help viewers decide if they should watch a video}
This summary is used as a cue for the viewer to decide whether to watch the video or not. To create a summary for this purpose, a system needs to extract scenes or frames which represent the content of the video well. Movie trailers and thumbnails are typical examples. For this kind of summary, attractiveness of the summary itself is also an important factor. When used in a video retrieval system, the relevance to the query is important.
\end{itemize}

Prior works explored various format of video summaries for these purposes.
In the following section, we overview the formats of video summaries.

\section{Output Format of Video Summarization}

The output format of video summarization can be categorized into static and dynamic summaries.

\subsection{Static summary}
We define static summary as a group of summary representation which does not change over time such as still images and text.
A typical example of static summary is keyframes which are frames presenting important content in a video. This format's advantage is that the summary itself can be viewed in a short time. Static summary is so flexible that we can layout static component to emphasize the structure of video content. We review some examples of static summaries.

\paragraph{Keyframes}
Keyframes are a set of frames from moments of interest as in \citep{Lee2012,Gong2000,Ma2002}. Keyframes are useful to quickly check visual contents of the video although audio and motions are lost in this format. Video thumbnail selection can be viewed as summarizing a video with keyframes \citep{liu15_cvpr,song2016click,yuan19_mm}. This is one of the simplest forms of a summary.

\begin{figure}
\begin{minipage}{.35\linewidth}
\centering
\includegraphics[width=\linewidth]{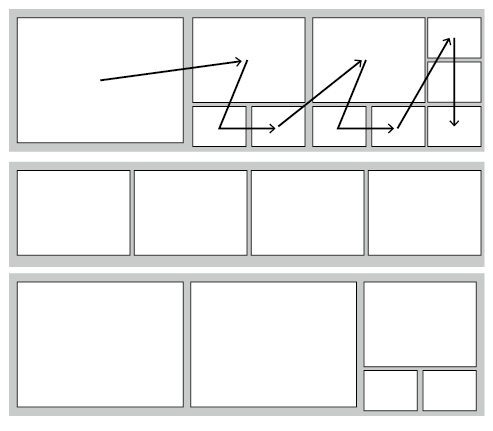}
(a) Comic book-style summary
\end{minipage}
\hfill
\begin{minipage}{0.65\linewidth}
\centering
\includegraphics[width=\linewidth]{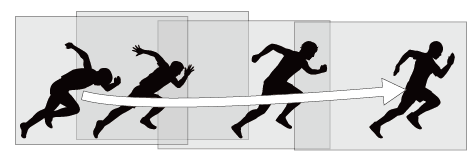}
(b) Storyboard-style summary
\end{minipage}
    \caption{Illustrations of static summary formats which display keyframes in special layouts. (a) In VideoManga \citep{Uchihashi1999}, keyframes are packed in a comic book style layout. Each row block (gray area) displays a sequence of keyframes from top to bottom and left to right. (b) Storyboard format represent a short scene with stitched keyframes and an arrow indicating the motion of a salient object \citep{Goldman2006}.}
    \label{fig:frame_composite}
\end{figure}

\paragraph{Composite of frames}
Several studies have attempted to express the structure of video content by presenting keyframes in a special layout (Fig.~\ref{fig:frame_composite}). VideoManga, for example, scales the keyframes and arranges them in a way that mimics a comics to visually express the relative importance of each moment \citep{Uchihashi1999}. Schematic storyboard proposes a summary that shows how a scene progresses by stitching frames and adding symbols that represent motion's direction \citep{Goldman2006}.

\paragraph{Other special formats}
Video summagator proposed a volume-based interface for video browsing \citep{Nguyen2012}. Such format is useful to understand spatio-temporal dynamics of a video and locate a moment of interest.
Multimodal representation has also been explored, \eg, pairs of a textual summary and a thumbnail to represent sections of a video \citep{mixt,videodigests}. The static components lead a user to quickly access the corresponding video chapter for more details.

\subsection{Dynamic Summary}
A dynamic summary is a short video summary of a video. A short video summary created by connecting the extracted segments is called a video skim. Since video skims can convey information with audio, video skims are suitable for summarizing videos such as lecture videos and movies where audio is important. Video skims are also suitable for summarizing sports videos because video skims show motions.
A single excerpt from a video can be viewed as a dynamic summary. Some video service providers, such as YouTube and Netflix, use a short preview of a scene as a dynamic thumbnail to help viewers to select whether they should watch the video.

\chapter{Video Summarization Approaches}
Video summarization problems, including video skimming, highlight detection, \etc, have different problem formulations, but also share the common goal of identifying content of interest from videos.
Previous research has proposed various approaches towards achieving this common goal. 
We will discuss approaches of three major categories.
First, we describe heuristic approaches. 
A typical heuristic approach uses domain-specific knowledge to create video summaries. 
Some methods use domain-specific external information other than videos. 
The literature in this category provides ideas useful for systems where an input video's domain and characteristics are known.
The next is learning-based approaches, which has been heavily researched in recent years for scenarios involving general videos with a sufficient amount of training data. 
We review unsupervised, supervised and weakly supervised approaches.
Finally, we describe methods for creating personalized summaries. 
This category is essential for practical systems because what is important about a video might depend on the viewer's preferences.

Before we explain each approach, we overview typical video summarization objectives.
This chapter introduces various methods, but we can interpret most of them as implementing some combinations of the following objectives.
\begin{itemize}

\item \textbf{Interestingness:} Frames or segments in summary videos should capture interesting events. For example, a moment of scoring in a sports video or a landmark scene in a tourism video would be necessary to comprise a good summary. Interestingness may depend on personal preferences, so some video summarization applications estimate what is interesting from user's preference. 

\item \textbf{Conciseness:} Repetition of similar scenes or long unchanging scenes should be omitted to summarize a video concisely. For instance, summarizing rushes from movie production needs to remove redundant retake scenes. The conciseness objective is employed by many unsupervised approaches.

\item \textbf{Representativeness:} The most representative segments or frames among candidates should be added to the summary. Important segments or frames can be accompanied by visually similar duplicates. To select one from such duplicates, segment or frame selection is optimized for representativeness objective.

\item \textbf{Diversity:} The content included in the summary should be diverse to capture the story of a video. This objective also encourages the summary to cover more content in the video.

\item \textbf{Coverage:} The summary should contain a complete set of content, preventing the loss of important content from the original video. Some methods trying to create a summary that reconstructs the original video with as small errors as possible. This approach is based on an assumption that efficiently sampling excerpts from the whole video can reconstruct remaining scenes with small errors. 
\end{itemize}

As we have discussed, the desired video summaries depend on the domain and the application scenario. 
Therefore, one must carefully consider what objectives to optimize when developing a video summarization system. 
For example, in a surveillance video summarization system, the coverage of anomaly events would be the top priority. 
For fixed-point videos streaming the wild life, we could create a good summary by removing redundant scenes and maximizing the diversity. Table~\ref{tab:objectives_1} and \ref{tab:objectives_2} summarizes objectives employed by existing video summarization methods.
\begin{center}
\begin{table}[h!]
\rowcolors{1}{}{lightgray}
    \centering
    \begin{tabular}{rccccc}
    \toprule
    & \rot{Interestingness} & \rot{Less-redundancy} & \rot{Representativeness} & \rot{Diversity} & \rot{Coverage}       \\ \midrule
    \cite{Smith-1997-14398}                & \checkmark & \checkmark & & & \\
    \cite{Ma2002}                          & \checkmark & & & & \\
    \cite{Laganiere2008}                   & \checkmark & \checkmark & & & \\
    \cite{10.1145/2601097.2601198}         & \checkmark &  & & &\checkmark \\
    \cite{Li2003AGF}                       & \checkmark & & & & \\
    \cite{10.1145/319463.319691}           & \checkmark & & & & \checkmark \\
    \cite{Babaguchi2004}                    & \checkmark & & & & \\
    \cite{Sang2010}                         & \checkmark & & & \checkmark & \checkmark \\
    \cite{Taskiran2006}                     & \checkmark & & & & \checkmark \\
    \cite{Aizawa2001}                      & \checkmark & & & & \\
    \cite{Sawahata2003}                    & \checkmark & & & & \\
    \cite{Gong2000}                        & & \checkmark & \checkmark & & \\
    \cite{Uchihashi1999}                   & \checkmark & \checkmark & \checkmark & & \\
    \cite{Daniel}                          & & & \checkmark & & \\
    \cite{1391003}                          &\checkmark & & & & \\
    \cite{6786125}                         & & \checkmark & \checkmark & & \\
    \cite{Zhao2014}                        & & \checkmark & & & \checkmark \\
    \cite{7934321}                         & & \checkmark & & & \checkmark \\
    \cite{Behrooz2017}                     & & & & \checkmark & \checkmark \\
    \cite{Rochan_2019_CVPR}              & & & & \checkmark & \checkmark \\ \bottomrule
\end{tabular}
\caption{Video summary objectives employed by each method}
\label{tab:objectives_1}
\end{table}
\end{center}

\begin{table}[h!]
\rowcolors{1}{}{lightgray}
    \centering
    \begin{tabular}{rccccc}
    \toprule
    & \rot{Interestingness} & \rot{Less-redundancy} & \rot{Representativeness} & \rot{Diversity} & \rot{Coverage}       \\ \midrule
    \cite{gygli14_eccv}                   & \checkmark & & & & \\
    \cite{zhang16_eccv}                   & \checkmark & & & \checkmark & \\
    \cite{7780481}                   & \checkmark & & & & \\
    \cite{10.1145/3123266.3123328}         & \checkmark & & & & \\
    \cite{Zhang_2018_ECCV}               & \checkmark & & & & \checkmark \\
    \cite{fajtl2018summarizing}            & \checkmark & & & & \\
    \cite{10.1145/3343031.3350992}         & \checkmark & & & \checkmark & \\
    \cite{10.1145/3240508.3240651}         & \checkmark & & & & \\
    \cite{062a3168836c4397a23d88be8139bd68}& \checkmark & & & \checkmark & \checkmark \\
    \cite{10.1007/978-3-030-01258-8_22}   & \checkmark & & & \checkmark & \\
    \cite{Liu_2020_ACCV}                 & \checkmark & & & & \checkmark \\
    \cite{narasimhan2021clip}                & \checkmark & & & \checkmark & \checkmark \\
    \cite{potapov14_eccv}                  & \checkmark & & & & \\
    \cite{Xiong_2019_CVPR}                & \checkmark & & & & \\
    \cite{Khosla2013}                      & \checkmark & & \checkmark & & \\
    \cite{Chu2015}                         & \checkmark & & & & \\ \bottomrule
\end{tabular}
\caption{Video summary objectives employed by each method}
\label{tab:objectives_2}
\end{table}

\section{Heuristic Approaches}
There are two groups in heuristic approaches.
The first develops heuristic rules based on low-level features for general video summarization.
For example, early work addresses video summarization by detecting visually stimulating signals or changes in low-level features.
The other group uses rules based on domain knowledge.

\subsection{Heuristic Approaches for General Video Summarization}
Heuristic rules to characterize important segments or frames are studied especially in the early days of video summarization.
In \citep{Smith-1997-14398}, the authors propose heuristic rules to select keyframe candidates based on observations from professional video editing patterns. 
For example, they observe that scenes in edited videos tend to contain faces and overlay text describing the scene.
Based on this, they compile heuristic rules to extract a frame with text when a scene contains both faces and text. 
Another rule is based on camera motion.
They propose to extract frames at the start or the end of zoom or pan motion. 
This rule is based on the observation that camera motions guide the viewer to the focal point in a scene. 
They propose eight heuristic rules and a final summary is created by combining those heuristic rules.
The method in \citep{Ma2002} uses a combination of three attention models to estimate which segments get the viewer's attention.
The attention models are based on several heuristic rules that involve visual and audio saliency, camera motion, and face positions.
For example, one heuristic is based on an intuition that a segment with closed-up faces likely gets human attention.
The method in \citep{Laganiere2008} assumes that salient objects or actions correspond to spatio-temporal changes in pixel values.
Based on this assumption, the method extracts interesting frames by detecting feature points with large variations of pixel values in spatial and temporal dimensions.
To this end, the method computes the Hessian matrix of spatio-temporal signals and the determinant of the Hessian which shows positive responses to blob-like features. 
\cite{10.1145/2601097.2601198} summarize videos that capture an event from multiple cameras into a single video. Their method is based on the observation that important content attracts more camera operators' attention. The method estimates camera motion and finds content captured by multiple cameras. When editing multiple videos into a single video summary, cinematographic guidelines are used to select a camera view, \eg, avoiding transitions from a camera to a camera with similar angles.

\subsection{Domain knowledge-based Approaches}
As there are many open problems in video recognition, domain knowledge-based techniques are promising. We revisit domain knowledge-based video summarization, which tackles the problem in various ways including using external sources of information.

One way to obtain domain knowledge is by understanding video content and how video is created in the production pipeline.
For example, sitcom TV shows have laughter sound inserted after each interesting moment; this kind of sound can be regarded as domain knowledge.
By detecting these patterns, we can make a summary consisting of humorous scenes.
Sports rules are also a type of domain knowledge useful for summarizing sports video.
In this sense, domain knowledge helps make a video summarization problem manageable. Recognizing events in a video and estimating the interestingness of the event is a challenging vision problem, but with the help of domain knowledge, the visual and audio understanding problem can be much easier, and sometimes one can even bypass the problem. For example, detecting an exciting play in a broadcast sports video seems to be a mixture of multiple vision problems that involve pose recognition, scene understanding, and possibly skill assessment. However, if we know that the broadcaster often highlights exciting plays with slow-motion replays, we can turn the event recognition and excitement estimation problem into replay detection \citep{Li2003AGF}. 

In certain domains, interesting events can be detected from low-level video features.
\citep{10.1145/319463.319691} use heuristic rules to detect interesting segments in lecture videos.
One heuristic is based on an observation that the pitch of speech is a cue for emphasis.
They compute ``pitch activity'' from the fundamental pitch frequency and add segments with high pitch activity to the summary.
Another heuristic is based on an assumption about slide transitions: a slide transition represents a change of topic and that the time spent on a slide represents the relative importance of that topic.
Based on this assumption, the method assigns time to each slide as a percentage of the time actually spent by the speaker.
If we can successfully create heuristic rules based on correct assumptions, a rule-based algorithm can reasonably solve the summarization problem.
One disadvantage is that such heuristic rules are less robust to common noises.
For example, the lecture video summarization may fail if there are unexpected audio noises such as audio content in slides or speech from the audience.

For some particular video domains, there exists domain-specific data that can be a good indexer of video content.
There are several studies that use domain-specific data for video summarization \citep{Babaguchi2004,Sang2010,Taskiran2006}.
\cite{Babaguchi2004} proposed an approach for American football videos that uses game statistics to detect interesting moments in videos.  
Game statistics provide time-stamped records of important events in a game as well as the players associated with those events, which we can obtain from sports news or team websites.
Aligning game statistics with the associated video allows the system to extract interesting events.
As this type of data is available in various sports, this approach could be applied to other broadcast sports as well.

For movies, \citep{Sang2010} use scripts for movie content analysis.
A script is a written description of movie content that includes characters, dialogues, and actions.
They find the most likely correspondences between a scene in a movie and one in the script by comparing histograms of characters.

For egocentric videos captured by wearable cameras, \citep{Aizawa2001} propose to monitor the brain waves of the person wearing the camera. Brain waves reflect the person's physiological status. The method tries to detect scenes in which the person felt interested by checking typical patterns found in brain waves. They further explore other data sources such as GPS, accelerometer, and gyroscopic sensors useful for understanding egocentric videos \citep{Sawahata2003}. For instance, when we want to search for a conversation scene in an egocentric video, we can filter out scenes where people move around and search for scenes of interest from the remaining parts.

\section{Machine learning-based Approaches}
\subsection{Unsupervised Approaches}
Unsupervised approaches are a promising option when domain knowledge and labeled summary examples are unavailable.
As they do not rely on domain knowledge or domain-specific datasets, they are intended to be for generic video summarization.
As the availability of sufficient ground truth summaries is unrealistic in most scenarios, substantial studies on machine learning-based video summarization focus on unsupervised method.

Clustering visually similar video frames or segments is a straightforward way to tackle video summarization by reducing redundancy and finding representative parts.
\citep{Gong2000} use singular value decomposition (SVD) to project frame features and cluster visually similar frames.
The frames closest to the cluster centers are extracted as keyframes so that the method prioritize representative frames.
For video skim generation, the longest shot of each cluster is selected.
Similarly, Video Manga \citep{Uchihashi1999} performs clustering of video frames using color histograms. They use frame clusters to segment a video and estimate the segments' importance scores. Specifically, a video is segmented based on which cluster each frame belongs to. A segment's importance score is computed based on the segment's length and the importance score of its associated cluster. The cluster importance is the relative duration that the cluster occupies in the video.
In \citep{Daniel}, frame clustering is formulated as a curve simplification problem to use the sequential structure of a video.
The method transforms a video into a trajectory curve of frame features, and the curve is divided by keypoints by the curve simplification algorithm.
The keypoints are regarded as a representative summary of the video.
In \citep{1391003}, clustering segments is achieved by graph partitioning, and the clusters are grouped into scenes by graph analysis. After scene detection, summarization is done in a top-down manner which extracts parts at scene, cluster, segment, and subsegment levels.
\citep{6786125} propose to use hierarchical clustering based on visual similarity and the sequential structure of a video.

Sparse coding has been used to tackle video summarization. 
Sparse coding learns a dictionary so that the data can be sparsely reconstructed using the learned dictionary.
The technique suits to some video summarization objectives, \ie, uniqueness, representativeness, and coverage.
In \citep{Zhao2014}, a video is processed from the beginning and the dictionary is gradually updated on-the-fly.
If the dictionary cannot reconstruct a newly passed segment at that point, the segment is added to a summary.
\citep{7934321} formulate the summary generation as a sparse coding problem and find a small subset of segments that can successfully reconstruct a multi-view video. The subset is regarded as an informative summary.

Some studies are based on an intuition that a good summary should contain minimal yet sufficient information for a good reconstruction of the original video.
In \citep{Behrooz2017}, an encoder-decoder model is trained so that the encoder selects frames and tries to reconstruct the summary into the input video.
The method further applies ideas from adversarial learning~\citep{goodfellow2014generative}: The frame selector and the decoder are trained to ``fool'' the discriminator, whose goal is to distinguish the reconstructed video from the input video.
Similarly, \citep{Rochan_2019_CVPR} use adversarial loss and reconstruction loss to train the keyframe selector.
In this method, unpaired data consisting of the raw video and real summaries are used for training.

\subsection{Supervised Approaches}
Supervised approaches is an option if a plenty of paired data of videos and their summary is available.
Taking video frames as input, the summary model estimates the frames' importance.
The outputs are compared to the ground truth annotation (importance scores) that indicates whether the frames should be added to the summary, and the estimation error is calculated.
In a supervised approach, the summary model is trained mainly by minimizing the estimation error.

One of the reasons for the recent rise of supervised approaches is the release of datasets tailored for video summarization.
In particular, SumMe~\citep{gygli14_eccv} and TVSum~\citep{Song2015} have had a significant impact on the development of machine learning-based video summarization approaches.
SumMe is a dataset with manually created video summaries for user videos.
In the paper where SumMe was introduced~\citep{gygli14_eccv}, the authors also proposed a supervised approach that learns to predict importance scores from video features such as faces, objects, and aesthetic features.
TVSum is another impactful video summary dataset whose videos are annotated with importance scores for every two seconds. The densely annotated importance scores are widely used to train importance estimation models.

To model complex temporal dependency in a video in an end-to-end manner, deep neural networks have been used to model the importance of frames or segments.
Inspired by the success of the LSTM sequential modeling in speech recognition and visual captioning, \citep{zhang16_eccv} use a model with an LSTM layer to predict the importance of video frames.
Taking a sequence of video frame features as input, their model is trained to produce an importance score for each time step.
To capture motion features of multiple frames, 3D convolutional neural networks are also used. A two-stream network that combines 2D features and 3D features of videos are proposed for highlight detection~\citep{7780481}.
By stacking LSTM layers, some work develop hierarchical LSTM models that aim to model intra- and inter-segment dependencies \citep{10.1145/3123266.3123328,Zhang_2018_ECCV}.
The first LSTM layer typically encodes frames in a segment into a segment-level feature, and the second LSTM layer takes the sequence of segment features as input.

As attention mechanisms became popular in vision modeling, various models with attention mechanisms for video summarization were proposed.
VASNet proposes a model that uses a self-attention mechanism to predict the importance of a frame \citep{fajtl2018summarizing}.
The self-attention mechanism is expected to reflect the context obtained from other frames.
In order to better model the long-term temporal dependency of video frames, memory networks are also considered \citep{10.1145/3343031.3350992,10.1145/3240508.3240651}.
Memory networks have a memory component that neural networks can read and write to for inference.
The purpose of using memory networks for video summarization is to store the visual information of the entire video and use it as context to predict the importance of frames or segments.
Graph convolutional networks (GCNs) have also been used for video summarization \citep{062a3168836c4397a23d88be8139bd68}.
The approach using GCNs focuses on modeling the relationships between frames. 
Nodes of a relation graph are the frames, and edges are the affinities between frames.
Another study formulates video summarization as 1D temporal segmentation, classifying each segment as a summary or background using a fully convolutional neural network \citep{10.1007/978-3-030-01258-8_22}.
Recent work uses transformers \citep{Liu_2020_ACCV,narasimhan2021clip}.
\citep{narasimhan2021clip} propose a multimodal transformer for query-guided summarization, which stacks a multimodal transformer and a frame-scoring transformer.
The multi-modal transformer extracts frame features guided by a query, and the frame-scoring transformer estimates frame importance scores from the outputs of the multi-modal transformer.

As we reviewed, the video summarization community actively explored a wide range of network architectures.
On the other hand, recent work that uses a simple model such as one fully-connected layer still shows competitive performance \citep{Saquil_2021_ICCV}.
Unfortunately, fair comparison of neural network architectures is a challenging problem. Training deep models is also costly and tuning hyperparameters sufficiently for many architectures is often infeasible. The optimization methods and regularization methods also affect the performance. Due to these challenges, much work is needed to understand what are the important components in a neural network architecture for video summarization.

\subsection{Weakly supervised Approaches}
Frame-level and segment-level annotations are costly, so real-world applications often face a shortage of training data. The goal of the weakly supervised approach is to facilitate training of video summarization models with coarse annotation data that is easy to collect. Existing work explores various annotations that can substitute densely annotated importance scores or ground truth summaries. Videos may be associated with metadata such as topics, durations, and titles. Such large-scale yet cheap video-level annotations are promising source of data to investigate importance of video content from another aspect.

Some work use video topics to infer the importance of segments. In such work, the summarization objective is to find content ``typical'' for a particular topic. For example, if the input video is a shot of a wedding reception, the video tends to include segments on the bride and groom, wedding cake, speeches, \etc. Topic-specific summarization assumes that such typical content is important and that other segments are irrelevant to the main content of the video. The method in \cite{potapov14_eccv} trains a support vector machine (SVM) that identifies whether a video belongs to a particular topic and then uses the SVM to infer the importance of a segment.

A study by \cite{Xiong_2019_CVPR} hypothesizes that the duration of a user-generated video is a good indicator of its importance. When users edit a video to make it shorter, they select segments that they think are the most important. Based on this insight, they train a ranking model that prioritizes segments likely to belong to shorter videos. Video duration does not need manual annotation effort, so the method enjoys large-scale training data. In the work, they collect user-generated videos on an social networks.

Some work address video summarization using weak supervision obtained from images or videos on the web. One advantage of web data is its scale. Web facilitates noisy yet large-scale data collection. Another advantage is that data on the web represent people's natural behavior. By investigating how people are engaged with web images and videos, we can find people's interests. For example, by querying Paris on the web, we would get pictures of landmarks like the Eiffel Tower, Arc de Triomphe, or Louvre. The top results represent what people are most interested in about Paris. The underlying idea of video summarization using web priors is to estimate content's importance based on people's interest in the web. The work in \citep{Khosla2013} summarizes videos capturing a particular object, \eg, car. Web images relevant to the category of interest are used to find canonical viewpoints of the object category. Segments that show the object from canonical viewpoints are added to the summary, and those of uninformative views are omitted.
\citep{Chu2015} assume that important content of an event frequently appears over videos of the same topic.
The authors collect videos on the same topic on the web to find such important content, and develop a method to discover visually similar segments that co-occur over a video collection.

Web images and videos are useful sources for obtaining priors on various topics, but there are several limitations. One is that the web changes over time, so relying on web images and videos can lead to unstable performance. Another limitation is that web biases can sneak into video image summarization algorithms. It is known that biases in data can make algorithms perform unfairly. For example, the study in \citep{DeVries2019} reports that the performance of popular image recognition services tend to degrade in non-Western regions or regions with lower incomes. Similarly, video summarization approaches using web priors may perform worse for user subgroups. There is a growing concern about biases in machine learning-based technologies \citep{Mehrabi2021}. It is important for researchers and developers to understand and identify possible problems when building machine learning-based video summarization systems.

\section{Personalizing Video Summaries}


Most work in video summarization focuses on learning a generic model that will be applied to all videos regardless of the intended viewers. However, different users often have different preferences in terms of what they consider to be good summaries even for the same input video. For example, let us consider a long video captured during a family vacation. The video may contain many different events and contents, such as food, sightseeing, activities, \etc. Suppose we would like to create a summary video for each family member, but different family members may have different preferences -- some may prefer seeing more content on food, while others may prefer content on sightseeing. Ideally, we would like the video summarization model to adapt to each person's reference and taste. The traditional approaches of learning generic video summarization models lack the ability to adapt the predictions to each user's preference.

To address this limitation, there has been some recent work on personalizing video summaries. The goal of this line of research is to develop video summarization models that can take advantage of users' preferences and adapt to each user accordingly. The user preference can be expressed in many different ways, including user history, query sentence, \etc. In this section, we review previous work on personalized video summarization.

\subsection{User Preference Representation}
Personalization requires user preference information. In the literature, modeling user preferences has been explored from different angles.

Some of the early work before the deep learning era capture user preferences by explicitly building or acquiring user profiles using various forms of metadata. \citep{agnihotri05_mir} perform user studies (\ie giving personality tests to users) to extract user personality traits and use them to personalize multimedia summarization. However, it is difficult to scale this approach to represent user preferences in practice, making it difficult to integrate it in a machine learning based system. It is also difficult to apply this approach to new users, since it is impossible to perform a user study during testing. The method described in \citep{agnihotri05_mir} is mostly a hand-designed system and is evaluated only on a small number of users. In \citep{jaimes02_icip}, the authors use a similar user-profile approach for adaptation, where the user profile describes the user's preferences or interests. Unfortunately, these user profiles are largely hand-engineered and there is no learning involved, making it difficult to apply it in practice. \citep{varini15_mm} consider egocentric video summarization for the cultural tour application. In order to personalize the video summary, they assume that the user directly provides his/her preferences (\eg ``art''). This user-provided preference information is combined with GPS tracking information and an external knowledge base (\ie DBpedia) to infer semantic classes that are most relevant to the user's interest. These semantic classes are then used to select the relevant shots in an input video to create a summary video tailored to the user. However, this approach requires the user to explicitly express his/her preference. In addition, it is specifically designed for the culture tour application and may not generalize to the wider scenarios of video summarization.

There is also work on inferring user preferences by analyzing users' viewing behaviors. \citep{babaguchi07_tmm} propose to build user profiles via implicit feedback from the user while they are using the system. This is similar to how internet search engines use the implicit feedback (\ie whether a user clicks a search result) to improve its search engine. The user profile is then used for video retrieval and summarization. Compared with acquiring user profiles via explicit user studies like \citep{agnihotri05_mir}, using implicit feedback for user profiling has the advantage that the system can personalize and adapt to a user without relying on an elaborate setup, \eg, a user study. It is also possible to adapt to new users. One disadvantage is that it requires significant user interactions before the model can build an accurate user profile for effective personalization. \citep{zen2016mouse} use mouse activity as an indicator of interestingness in videos. They demonstrate their approach on a large-scale experiment involving 100K user sessions from a popular video website. In \citep{peng11_tmm}, the authors propose to analyze the user's viewing interest based on attention (\eg by analyzing eye movement, blink, head motion) and emotion (\eg by analyzing facial expression). The user's viewing behaviour is then used to personalize home video summarization. However, it relies on several auxiliary computer vision-based prediction components (\eg face/eye detection, head motion detection, blink/saccade detection, \etc.) which may introduce noise to the system arising from prediction failures. In addition, the approach assumes access to a camera that records users' viewing behaviours during deployment. This assumption may not be realistic in many real-world applications. 

Some work infers user preference by using an interactive approach. \citep{delmolino17_aaai} obtain user preference by asking questions about segments of a video such as ``Would you want this segment to be in the final summary?'' and ``Would you want to include similar segments?'' The user feedback is then used to refine the summary in an online fashion. This process is repeated until the user is satisfied with the result. The authors propose to infer both the personalized summary and the next question to ask with the goal of reducing the user interaction time needed to produce a good summary. \citep{jin2017elasticplay} propose ElasticPlay that allows users to interactively specify the remaining time budget given the generated summary so far. \citep{singla16_aaai} propose an interactive approach for image collection summarization by leveraging the user feedback in the forms of clicks or ratings.

Some recent work proposes to exploit user history to personalize video highlight detection. In \citep{delmolino18_mm,rochan20_eccv}, the authors consider highlight detection for automatic GIF creation by analyzing the user history, which is in the form of GIFs a user previously created. The assumption is that a user's GIF history provides fine-grained understanding of their interest. For example, a user's GIF history provides information about whether or not the user is interested in basketball videos, and if the user is interested in particular parts of basketball videos. Such fine-grained user preference information is often difficult to obtain from other metadata (\eg age, gender, interests in certain topics) previously used in highlight detection. The advantage of using user GIF history is that the model does not need to explicitly interact with users in order to acquire the user preferences. One disadvantage, however, is that this approach can only be applied if the users have created some GIF before, \ie it cannot be used for new users without GIF history. In addition, these approaches may not work well if the users are interested in diverse topics, \ie the GIF history covers different topics.

Another line of work uses text queries provided by users to represent user preferences for personalization in various video-related tasks. \citep{yang03_mm} proposes a video question answering system is developed for personalized news video retrieval. The user provides an input query in the form of a short natural language question. The system then retrieves the relevant videos and returns short precise news video summaries as answers. \citep{liu15_cvpr} propose a multi-task embedding approach for video thumbnail selection with side semantic information. The key observation is that online videos often come with side information (\eg title, description, query, transcript) This work proposes to combine this side semantic information with video content for thumbnail selection. In \citep{sharghi16_eccv,sharghi17_cvpr,vasudevan17_mm}, the authors develop query-focused extractive video summarization. Given an input video and a user query (\eg a list of object names or a short text phrase), the model returns a summary video by selecting key shots from the video according to the query, while maintaining the diversity of the selected shots. In \citep{Song2015,yuan19_mm,rochan20_bmvc,lei21_arxiv}, the authors consider the task of video highlight detection using language queries. Given an input video and a natural sentence query, these methods try to find the highlight (\ie moment) in the video according to the sentence query.

\subsection{Personalization Techniques}
Early work in personalized video summarization is mostly based on heuristic rules. However, it is difficult to develop a heuristic-based approach that generalize to the variety of real-world applications. In recent years, most research has been focused on methods using machine learning. In this section, we review on learning-based methods for personalization in video summarization.


\noindent {\bf User-specific Training:} One approach for personalization is to train a separate model for each person. For example, \citep{jaimes02_icip} propose a system for generating personalized video digests from MPEG-7 metadata. The system extracts feature vectors from existing metadata signal in a video. A model is then trained to classify each segment in a video as important/unimportant events using standard supervised learning. To achieve personalization, a separate model is trained for each person using training data specific to that person. Although this approach is simple and intuitive, collecting training data for each target user can be cumbersome and sometimes infeasible. 

\noindent {\bf Feature Fusion:} Another popular approach for personalization is to treat user preference information (\eg user profile, user history, text query, \etc) as additional features and fuse/concatenate it with visual features extracted from an input video. For example, \citep{delmolino18_mm} consider personalized video highlight detection for the application of automatic GIF creation from videos. They assumes that the user history is in the form of previously created GIFs denoted as $\mathcal{G}$. The proposed method learns a function $h(s,\mathcal{G})$ where $s$ is a candidate segment in a video and $h(s,\mathcal{G})$ is the score of the segment $s$ being the highlight. Note that $h(s,\mathcal{G})$ depends on the user history $\mathcal{G}$, so it is personalized to each user. The authors propose several different ways to fuse the user history $\mathcal{G}$ with the video segment $s$. For example, one way is to aggregate the GIF representation across all GIFs in the history to obtain an aggregated history $p$. The aggregated history $p$ is then directly concatenated with the feature representation of the segment $s$. Finally, a classifier can be learned based on the combined feature representation. In \citep{lei21_arxiv}, the authors consider the video highlight detection via natural language queries. The proposed method concatenates the projected video and text query features. The concatenated features are then used as input to a transformer-based encoder-decoder model for highlight detection.

\noindent {\bf Multi-modal Interaction:} One possible limitation of naive feature fusion -- \ie, direct concatenation of visual features and user preference features -- is that it does not sufficiently capture complex interaction between different modalities. To address this, some work builds models that capture the complex multi-modal interaction between the visual information and the user preference information. For example, \citep{yuan19_mm} consider the problem of sentence-specified dynamic video thumbnail generation. Given an input video and a query sentence, the goal is to generate a video thumbnail by selecting keyframes in the video. The generated thumbnail should match the query sentence. Personalization is achieved by allowing users to provide the sentence query as part of the input to the system. The method captures fine-grained interaction between sentence and video using a soft attention mechanism. Specifically, it calculates a soft attention score $\beta_n^t$ between the $t$-clip in the video and the $n$-th word in the query sentence. This soft attention allows the model to exploit the detailed interaction between clips in a video and words in a sentence. 

\begin{figure}
    \centering
    \includegraphics[width=0.5\linewidth]{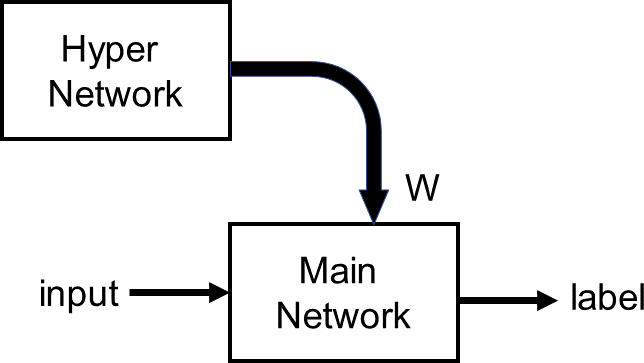}
    \caption{Illustration of the hypernet architecture. The main network (with parameters $W$) maps some input to its corresponding label. Different from standard neural networks, the parameters $W$ of the main network are directed predicted by another neural network (called ``hyper network'').}
    \label{fig:hypernet}
\end{figure}

\noindent {\bf Hypernetwork-based Model:} The idea of hypernetworks~\citep{ha17_iclr} is to use one neural network (known as a ``hypernetwork'') to generate the weights for another neural network (known as a main network). Similar to usual neural networks, the behavior of the main network is to map input signals to their corresponding labels. However, the behaviour of the hypernetwork is different from the usual neural network in the sense that it outputs the weight of another neural network (\ie the main network), see Fig.~\ref{fig:hypernet} for an illustration.

There is some recent work on personalization using ideas similar to hypernetworks. \citep{rochan20_eccv} propose a method for personalized highlight detection using user history. The model consists of two networks: a highlight detection network and a history encoder network. Using the terminology of hypernetworks, the highlight detection network is the main network and the history encoder network is the hypernetwork. The goal of the highlight detection network is to predict the target (\ie highlight) given the raw input (\ie raw video). The goal of the history encoder network is to take the user history as input and produce (a subset of) parameters for the highlight detection network. Specifically, the output from the history encoder network corresponds to the batch norm parameters in certain layers of the highlight detection network. For two different users, the output from the history encoder network will be different since the input (\ie user history) to the history encoder network is different. This means the highlight detection network will be different for these two different users because some of the network parameters (in particular, the batchnorm parameters) in the highlight detection network correspond to the output of the history encoder network. This is how personalization is achieved in the end. In \citep{rochan20_bmvc}, a similar approach is used for sentence guided video thumbnail generation. Given an input video and a sentence query provided by the user, the goal is to generate a video thumbnail that provides the preview of the input video. The generated video thumbnail should also be semantically relevant to the sentence query. In this work, a similar hypernetwork architecture is used. The proposed model consists of a thumbnail generation network (main network) and a sentence encoder network (hypernetwork). The sentence encoder network takes the sentence query as input and directly produces some model parameters in the thumbnail generation network.

\chapter{Benchmarks and Evaluation}
\section{Dataset}
Datasets are a critical component for developing and evaluating video summarization approaches.
Building a dataset for video summarization is not a straightforward task.
Prior research has explored what data should be annotated and how the data should be collected.

Some of existing video summary datasets provide pairs of videos and reference summaries \citep{gygli14_eccv,DeAvila2011,fu2017highlight}.
An early video summarization dataset, VSUMM \citep{DeAvila2011}, collected manually selected keyframes of videos from Open Video Project and websites including YouTube.
SumMe \citep{gygli14_eccv} is a widely used dataset for evaluating video summarization algorithms, which contains
25 personal videos collected from YouTube and the corresponding reference summaries.
As video summarization is highly subjective and there can be multiple ``good'' summaries for a single video, multiple reference summaries are usually collected for each video.

Instead of reference summaries, some datasets provide temporally annotated importance scores or ratings.
MED Summaries dataset \citep{potapov14_eccv} assigns category-specific importance scores to segments.
It provides category-specific importance scores obtained by asking annotators if each segment contains evidence that identifies event category of the video.
TVSum \citep{Song2015} provides importance scores annotated for every two second of each video.
They point out \textit{chronological bias} that segments appear earlier are likely to be rated higher.
To mitigate this bias, video segments are randomly reordered when they are displayed to annotators.
One advantage of collecting importance scores instead of reference summaries is that the duration of summaries does not have to be determined beforehand.
This allows more flexible evaluation.
During evaluation, reference summaries are created from importance scores using controllable settings, such as segment boundaries and the maximum duration of a summary.
A summarization system is evaluated by comparing its output summary and the reference summaries.

Some datasets provide textual annotations, \eg, concept labels and descriptions of video segments.
Textual annotations are easier to collect than manual summaries or temporally annotated scores.
\cite{Yeung2014} construct a egocentric video dataset VideoSET for video summary evaluation, where every segment is annotated with a textual description.
As substitutes for reference summaries, VideoSET provides textual summaries of the entire video content.
For evaluation, a video summary is converted into textual representation by retrieving textual descriptions associated with segments in the summary.
The quality of a summary is computed by comparing the textual representation and reference textual summaries.
This approach aims to assess the semantic similarity between created summaries and reference textual summaries.
VISIOCITY by \cite{Kaushal2021} is another dataset in this direction.
Videos in VISIOCITY are annotated with visual concepts including action, entity, scene, \etc.
The evaluation protocol by VISIOCITY uses a suite of metrics based on the textual annotations.
For example, the ``diversity'' metric is computed based on the overlap of concepts in selected segments.

Although video summarization datasets plays critical role, there are still some challenges in data for video summarization task.
Due to the high cost of human annotation, it is difficult to scale video summarization dataset.
The number of videos in most video summarization datasets are less than 200, which may be insufficient for recent deep learning based approaches that are data hungry.
Some work have crawled large-scale videos on the web \citep{Chu2015,gygli16_cvpr,Xiong_2019_CVPR}.
Such videos are often not fully annotated, but can be used for weakly supervised or unsupervised approaches.
Some recent surveys provide lists of existing video datasets \citep{9594911,Tiwari2021}.
\section{Evaluation Measures}
Evaluation methods are critical to understand the efficacy of video summarization systems and to measure the advances of new methods from existing ones.
However, evaluating video summary is not straightforward.
Different from other vision tasks, such as image classification or object detection, the existence of ground truth summaries is not clear.
One video can result in different video summaries which are all acceptable, and it is infeasible to list all possible summaries. 
We review some major ideas of existing evaluation methods.

\subsection{Manual Evaluation}
Manual evaluation is widely employed especially by early work in this area.
User study is a simple yet useful option to evaluate created video summaries.
In a standard user study, users are displayed with video summaries and are asked to rate the quality of the summary.
User study enables investigating complex property of summaries which are difficult to quantify with objective measures, \eg, aesthetics, and comprehensibility of a story.
The TRECVid video summarization task manually judges the quality of video summaries \citep{2020trecvidawad}.
In-house annotators watch submitted video summaries and their level of understanding is checked by asking questions about the content.
This evaluation protocol is based on an assumption that a good summary should convey the story of the original video while discarding less-important scenes.
Although a user study is able to investigate important properties of video summaries, setting up a user study is costly and difficult. Also, evaluation is difficult to reproduce and has a problem especially in the comparison of different systems.

\subsection{Quantitative Evaluation}
Quantitative evaluation have become popular in the recent years due to the convenience of not having to setup laborious user studies. It is also because the research community is increasingly more focusing on quantitative comparisons with existing approaches.
A common strategy for quantitative evaluation is to compare created summaries to reference summaries.
VSUMM \citep{DeAvila2011} collect reference keyframes by multiple annotators for each video and propose to evaluate keyframe extraction by comparing system-generated keyframes to reference keyframes.
The quality of automatic keyframe extraction is evaluated by computing the visual similarity to manually selected keyframes.

\paragraph{F1-score}
F1-score is a commonly used metric for video summarization to measure the similarity between generated and reference summaries.
F1-score is a precision and recall-based measure computed as:
\begin{eqnarray}
    \textrm{F1-score} &=& 2 \times \frac{\textrm{Precision}\times\textrm{Recall}}{\textrm{Precision}+\textrm{Recall}} \\
    \textrm{Precision} &=& \frac{\textrm{\# retrieved and relevant frames}}{\textrm{\# retrieved frames}} \\
    \textrm{Recall} &=& \frac{\textrm{\# retrieved and relevant frames}}{\textrm{\# relevant frames}}.
\end{eqnarray}
Specifically, retrieved frames are frames selected for a summary by a system, and relevant frames are those in a reference summaries.
The pairwise F1-scores are aggregated by taking an average or the maximum.
Taking the maximum value is based on an assumption that there can be more than one correct summary for a video, and that a summary similar to at least one reference is considered as a good summary.
Most conventional work measure an average score for the TVSum dataset~\cite{Song2015} and the maximum score for the SumMe dataset~\cite{gygli14_eccv}.
It is important to note that an average and the maximum values are not comparable to each other. Therefore, how the final performance score is computed needs to be clarified.
F1-score is the most common measure for video summary evaluation, however, \cite{otani2018} and \cite{10.1145/3394171.3413632} pointed out problems of this measure which will be discussed in Sec.~\ref{sec:eval_limitation}.

\paragraph{Experimental Setting}
For performing evaluation using the F1-score, there is a evaluation protocol introduced by \cite{zhang16_eccv} that is widely employed by recent papers on machine learning-based video summarization.
They propose to combines several datasets; SumMe, TVSum, Open Video Project (OVP) \citep{DeAvila2011,OPV}, and YouTube dataset \citep{DeAvila2011}, and evaluate models with three different settings.

\begin{itemize}
\item \textbf{Canonical:} The setting uses 80\% of SumMe for training and validation, and the remaining 20\% for testing. TVSum is also divided in the same way.
\item \textbf{Augmented:} The training set of ``Canonical'' is augmented with other datasets. For instance, 80\% of SumMe and the whole datasets of TVSum, OVP, and YouTube are used for training and validation; 20\% of SumMe is used for testing.
\item \textbf{Transfer:} The aim of this setting is to check if the model generalize to an unseen dataset. The whole datasets of TVSum, OVP, YouTube are used for training and validation, then, the whole SumMe is used for testing. When the model is tested on TVSum, the whole SumMe is used for training and validation.
\end{itemize}

Running experiments several times on random splits and reporting the average of the scores is a common practice.

\paragraph{Rank Correlation}        
Another approach for computing similarity to human reference is the use of rank correlation.
This measure computes the rank correlation between predicted frame importance scores and manually annotated importance scores.
It assesses intermediate output before post-processing to create output summary video.
Therefore, this measure can be used along with other complementary measures which evaluate final output summary.

\paragraph{Evaluation using Textual Annotation}
Some benchmarks with textual annotation employ their own objective measures.
VideoSET uses textual summary of a video content by annotators for evaluation \citep{Yeung2014}. 
A created video summary is converted into a textual representation by retrieving video descriptions associated with segments in the summary.
The similarity between the textual representation and the reference textual summary is computed using ROUGE, an NLP-based similarity measure for text summarization.
VISIOCITY evaluates several properties of a video summary using concept labels \citep{Kaushal2021}.
They design several metrics for the benchmark including content diversity, event continuity, and domain-specific importance.

\section{Limitations of Evaluation}
\label{sec:eval_limitation}
Evaluation protocols using the F1-score on public datasets have had great impact on video summarization as it allows researchers to compare a broad range of video summarization systems.
The standard protocol is to compute the F1-score on TVSum and SumMe datasets, and many machine learning-based methods use $k$-fold cross validation.
After the evaluation protocol is established, most works are evaluated using this protocol. However, some problems are revealed by recent analyses.

\paragraph{Post processing has dominant effects on F1-scores on TVSum}
The report by \cite{otani2018} demonstrates that the F1-score on TVSum is mostly determined by post processing, \ie, shot temporal segmentation and shot selection.
To evaluate a system on TVSum, reference summaries need to be created from frame importance scores.
The reference summary creation involves temporal segmentation and segment selection by a knapsack algorithm, which is biased to select short segments.
They empirically show that using the same temporal boundaries and knapsack algorithms are highly likely to output similar summaries to reference summaries, and results in high F1-score.
Therefore, the quality of importance score prediction is hardly reflected in the F1-score.
This also means that the measure can be ``gamed'' by selecting certain temporal segmentation methods.

\paragraph{Data splits make significant difference in F1-score}
\cite{10.1145/3394171.3413632} demonstrated that the performance is significantly affected by how the dataset is split into training and test splits.
This insight suggests that the comparison based on the reported scores using different data splits are not reliable.
They re-evaluated multiple video summarization methods on a large set of randomly created data splits and observed that the level of difficulty significantly varies over the set of splits, causing large variability in the F1-score.
As different random data splits are used for evaluation in each paper, reliability of the comparison based on the reported scores is limited.
They propose an evaluation protocol to mitigate the effects of data splits.
In the proposed protocol, the relative improvement from random summaries is computed on each data split so that the difference of difficulty is normalized.
\chapter{Open Challenges}

As we reviewed, there have been a wide variety of studies on video summarization exploring summary formats, applications, summarization methods, and evaluations. On the other hand, some large areas are still to be pursued. We believe that the following issues are critical for future research in video summarization.

\paragraph{Large-scale datasets with clean labels}
Large datasets with clean labels play an important role across the computer vision community. Without high-quality public datasets like ImageNet, MSCOCO, and Kinetics, various vision tasks would not be as successful as today. On the other hand, video summarization does not have such large-scale datasets, mainly because of the massive cost of building and annotating video summarization datasets. Creating reference summaries or densely annotating videos is labor-intensive and difficult to scale. Also, summarizing videos is a subjective task, and human annotators may have different opinions on what constitutes a good summary. Therefore, controlling the quality of labeling is also a challenge.

\paragraph{Difficulty of defining appropriate evaluation metrics}
Researchers have not established a way to evaluate the quality of video summaries. The quality of a video summary is highly subjective, and there can be multiple good summaries for one video. Moreover, desired properties vary depending on applications and video domains. Due to the lack of reliable evaluation methods, comparing video summarization methods is difficult. Similar to other creative tasks such as image generation, quantitatively evaluating video aesthetics is also an important issue but is hardly studied in this area.

\paragraph{Video summarization for emerging applications}
Technologies and cultures surrounding videos are changing day by day. For example, video conferencing has drastically spread in 2020 due to the global pandemic. Vlogging is now more popular than ever and short video platforms have acquired an enormous set of users who create and watch videos every day. Videos are also getting attention as a digital marketing tool, and a vast number of video ads are distributed on the web. Developing video summarization methods for emerging applications will stay an important area with new video tools and activities.
\chapter{Conclusion}
In this article, we provided an overview of various topics related to video summarization.
We described different types of videos and how their characteristics influence the desired format of video summaries.
We also reviewed a wide range of techniques and evaluation methodologies in video summarization, and 
discussed open challenges in this area.

Prior research on video summarization has considered diverse problems and explored various techniques.
However, video summarization is still challenging, and along with the change of technologies and cultures surrounding videos, new problems for video summarization emerges.
As creating and watching videos is more active than ever, video summarization now has the potential to have a larger impact on real-world services and people's daily lives.
We hope that this review of prior research will provide a starting point for the reader to analyze the problems and possible solutions.

\backmatter  
\printbibliography
\end{document}